\documentclass{article}

\usepackage[nonatbib, final]{nips_2017}

\usepackage[utf8]{inputenc} 
\usepackage[T1]{fontenc}    
\usepackage{hyperref}       
\usepackage{url}            
\usepackage{booktabs}       
\usepackage{amsfonts}       
\usepackage{nicefrac}       
\usepackage{microtype}      
\usepackage{color}
\usepackage{adjustbox}
\usepackage{tabularx}
\usepackage{amsmath,amssymb}
\usepackage{graphicx} 
\usepackage{subcaption} 
\usepackage{changepage}
\usepackage{float}

\usepackage[pdftex,dvipsnames]{xcolor}
\usepackage{xargs}
\usepackage[colorinlistoftodos,prependcaption,textsize=tiny]{todonotes}

\title{Learning Robust Dialog Policies in Noisy Environments}

\author{Maryam Fazel-Zarandi\\
Amazon Inc, USA\\
\texttt{fazelzar@amazon.com}\\
\And
Shang-Wen Li \\
Amazon Inc, USA \\
\texttt{shangwel@amazon.com} \\
\And
Jin Cao \\
Amazon Inc, USA \\
\texttt{jincao@amazon.com} \\
\And
\hspace{10pt}Jared Casale  \\
\hspace{10pt}Amazon Inc, USA \\
\hspace{10pt}\texttt{casjared@amazon.com} \\
\And
\hspace{11pt}Peter Henderson \\
\hspace{11pt}Amazon Inc, USA \\
\hspace{11pt}\texttt{pehender@amazon.com} \\
\And
David Whitney \\
Brown University \\
\texttt{dwhitney@cs.brown.edu} \\
\And
\hspace{7pt}Alborz Geramifard \\
\hspace{7pt}Amazon Inc, USA \\
\hspace{7pt}\texttt{alborzg@amazon.com}}

\begin{document}

\maketitle

\begin{abstract}
Modern virtual personal assistants provide a convenient interface for completing daily tasks via voice commands. An important consideration for these assistants is the ability to recover from automatic speech recognition (ASR) and natural language understanding (NLU) errors. In this paper, we focus on learning robust dialog policies to recover from these errors. To this end, we develop a user simulator which interacts with the assistant through voice commands in realistic scenarios with noisy audio, and use it to learn dialog policies through deep reinforcement learning. We show that dialogs generated by our simulator are indistinguishable from human generated dialogs, as determined by human evaluators. Furthermore, preliminary experimental results show that the learned policies in noisy environments achieve the same execution success rate with fewer dialog turns compared to fixed rule-based policies.
\end{abstract}

\section{Introduction}

Modern speech-based assistants, such as Amazon Alexa, Apple Siri, and Google Assistant, enable users to complete daily tasks such as shopping, setting reminders, and getting answers to factual questions using vocal commands. Such human-like interfaces create a rich experience for users by enabling them to complete many tasks hands-free and eyes-free in a conversational manner. Depending on how complicated a task is, multiple rounds of conversation may be needed for the assistant to fully understand user requests. To achieve this, these interfaces require the design of complex dialog policies which can generate appropriate responses to queries and steer the conversation. 

Designing dialog policies in current speech-based assistants is challenging and time-consuming. In particular, responding to a user can be difficult since the selection of system actions is conditioned on the task, the user goals and preferences, and the dialog history. Additionally, environment noise and ambiguous user utterances present added complexity. Communicating with speech-based assistants always involves a noisy communication channel and the assistant should be able to recover from automatic speech recognition (ASR) and natural language understanding (NLU) errors by confirming its recognition results or eliciting more information from the user. To address these issues, both rule-based strategies and statistical modeling techniques have been developed \cite{s_young_simulator}. Rule-based systems are costly to design and maintain, and generally assume the dialog state is fully observable with only a limited account of errors and uncertainties. Statistical modeling techniques, on the other hand, are able to generalize to such unknowns, but rely on the availability of huge amounts of training data. Some statistical-modeling techniques formulate dialog management as a reinforcement learning (RL) problem \cite{levin}. In this formulation, the goal is to adjust a parameterized dialog policy in order to maximize the expected cumulative reward over the course of a dialog. To reduce the cost of collecting the large amount of data needed by such models, researchers have focused on the development of user simulators which interact with the dialog policy (e.g., \cite{simulator_hmm, user_modeling_simulator, maluuba, georgila, s_young_simulator, s_young_simulator_2}). Most of the existing simulators are domain dependent \cite{maluuba} and lack robustness against noise in real user utterances in order to recover from ASR/NLU errors \cite{neural_dialog_system}.

In this paper, we focus on learning dialog policies in noisy environments where the assistant needs to recover from ASR/NLU errors. The key contribution of this paper is the development of a realistic user simulator which interacts with the assistant through voice commands. The simulator is trained on real-world data and takes dialog context into account. Furthermore, while other works simulate NLU errors at the intent- or slot-level~\cite{simulator_vivian}, our simulator is able to use artificial noise in audio signals to induce a similar corpus-specific ASR word-error rate (WER). This error can then propagate to realistic intent- or slot-level errors. Using dialogs collected from MovieBot \cite{MovieBot}, an Alexa Skill \cite{ask} that converses with users about existing, new, and forthcoming movies, we show that our simulator generates dialogs that are indistinguishable from human generated dialogs. Using the simulator, we leverage deep RL to learn robust dialog policies that overcome upstream ASR/NLU errors in a noisy environment. Experimental results show that the learned policies achieve the same execution success rate with fewer dialog turns compared to fixed rule-based policies in noisy environments.

\section{Related Work}

A dialog system can be formalized as a Markov Decision Process (MDP) \cite{levin}. An MDP is a tuple <$S, A, P, R, \gamma$> of states, actions, transition probability function, reward, and discount factor. In the formulation of RL, at each time step \textit{t}, the agent observes a state $s_t \in \mathcal{S}$ and selects an action $a_t \in \mathcal{A}$ according to its policy ($\pi: \mathcal{S} \rightarrow \mathcal{A}$). After performing the selected action, the agent receives the next state \textit{$s_{t+1}$} and a scalar reward \textit{$r_t$}. The trajectory restarts after the agent reaches a terminal state. With this formalization, RL can be used to find the optimal dialog policy (e.g., \cite{singh, pomdp_dm, lets_go, negotiation}). In this context, at each turn the system acts based on its understanding of what the user has said, and the reward function is modeled in terms of different dimensions such as per-interaction user satisfaction, accomplishment of the task, efficiency of interaction, dialog duration, etc. 

Recently, deep RL has also been applied to the problem of dialog management. Cuayahuitl \cite{simple_ds} presented an implementation of DQN \cite{deep_rl, deep_rl_2}. Zhao and Eskenazi \cite{drl_with_state} similarly used a deep recurrent Q-network (DRQN) to learn both state tracking and dialog policy for task-oriented dialogs. Fatemi et al. \cite{fatemi} compared different deep RL techniques on a restaurant-finding dialog problem. They showed that DDQN \cite{vanhasselt} converges faster than DQN for this domain.

An important challenge in using RL for learning dialog policies is creating realistic user simulators that can generate natural conversations similar to a human user \cite{s_young_simulator}. Existing approaches for simulating users can be categorized into two groups: intent-level and utterance-level simulation. An intent-level simulator abstracts a conversation as a sequence alternating between user intents and slots and bot actions, where an intent indicates the user's intention (e.g., getting the plot) and slots represent information about a particular entity (e.g., movie title). The goal is to predict the next user intent based on the dialog history (e.g., \cite{user_modeling_simulator, simulator_longer_context, simulator_vivian, agenda_simulator}). These methods require extensive work to manually design the user goal or agenda, which makes them less generalizable. Additionally, it is difficult to simulate real-world recognition errors. Utterance-level user simulators, on the other hand, attempt to generate natural language user responses. To this end, template-based natural language generation is usually adopted to output utterances based on handcrafted rules and predicted user intents \cite{simulator_vivian}. Inspired by recent studies in sequence-to-sequence machine translation \cite{seq_2_seq_mt}, end-to-end dialog simulators generate utterances with minimal feature engineering \cite{maluuba, crook}. However, these techniques often produce generic responses \cite{deep_rl_dg}. Additionally, they have to be trained with a significant amount of dialogs, and thus they do not work well for domain-specific applications.

Given that the dialog state is not fully observable due to ASR/NLU errors and uncertainty over a user's goals, researchers have also investigated the use of Partially Observable Markov Decision Processes (POMDPs) \cite{pomdp}. However, these systems are complex and exact policy learning for POMDPs is intractable \cite{pomdp}. Rule-based policies, on the other hand, use ASR/NLU confidence scores to recover from errors. However, asking too many clarifying questions can frustrate users and in some scenarios it might be more beneficial to execute on available information. Additionally, ASR/NLU components are not static and as such policies with fixed confidence levels need to be constantly updated.

\section{User Simulation}

\begin{figure}[b]
  \centering
  \includegraphics[width=0.8\textwidth]{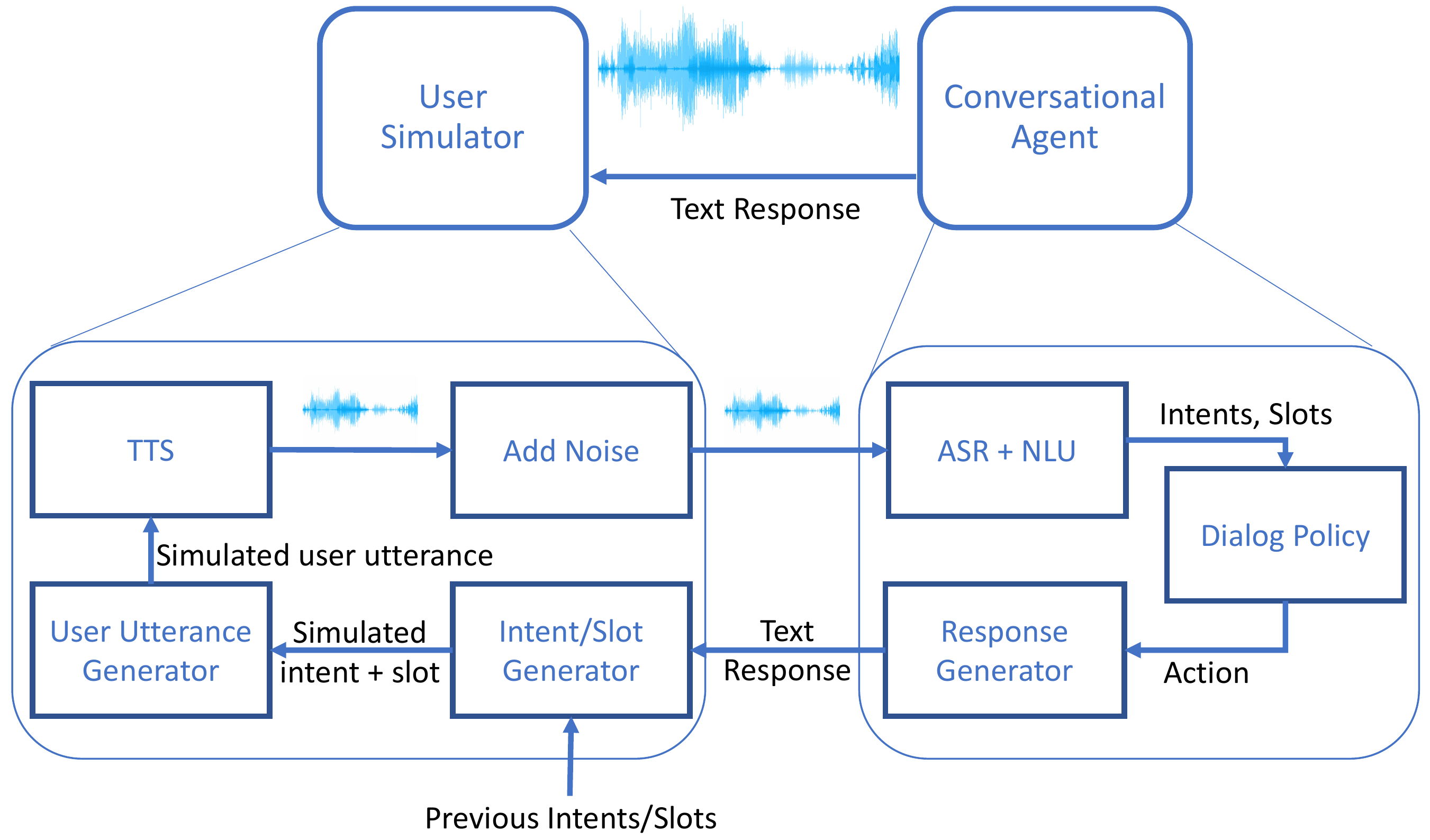}
  \caption{User simulator and conversational agent interaction. We use the text response generated by the agent before sending it to TTS to eliminate the need for ASR/NLU on the user simulator side.}
  \label{simulator-diagram}
\end{figure}

To develop a user simulator that generates voice responses to the agent, we adopt a hybrid of language model based intent prediction and template-based utterance generation. The overall architecture of the user simulator is illustrated in Figure \ref{simulator-diagram}. 

The user simulator first generates the next intent and slot type based on dialog context. Dialog context includes previous intent and slot type pairs in the conversation, as well as text responses from the agent encoded with word2vec \cite{word_to_vec}. We model intent generation as a language modeling problem and use recurrent neural networks (RNNs) to predict the trajectory of intents. In this formulation, each possible intent and slot type pair form a token in the vocabulary, and every training dialog becomes a training intent sequence. For example, the sequence for the conversation with MovieBot in Table \ref{moviebot-example} is \textit{[Start, GetGenreMoviesIntent+genre, GetNextActorIntent, GetRatingIntent, End]}. 
 
Given the predicted intent and slot type, the user simulator uniformly samples one utterance from the templates and fills the utterance with slot values uniformly sampled from a lexicon of the corresponding slot type. For example, given intent \textit{GetDirectorIntent} and slot type \textit{movie\_title}, a sample utterance from the template is "\textit{who is the director of $\left\{movie\_title\right\}$}?". $\left\{movie\_title\right\}$ is then replaced uniformly by a value from the lexicon of movie titles. When the simulator is responding to a clarifying question, it answers \texttt{yes} or \texttt{no} to a confirmation question, but repeats the original utterance for elicitation. 

\begin{table}[t]
 \small
  \centering
  \caption{Example of a conversation with MovieBot. The intent indicates the user's intention and slots (e.g., genre) represent information about a particular entity.}
  \label{moviebot-example}
  \centering
  \begin{adjustbox}{max width=\textwidth}
  \bgroup
  \def\arraystretch{1.1}
  \begin{tabularx}{\linewidth}{lX}
    \toprule
    \multicolumn{1}{c}{\textbf{Speaker}} & \multicolumn{1}{c}{\textbf{Utterance}} \\
    \midrule
    User & A science fiction movie [intent = GetGenreMoviesIntent, genre = science fiction] \\
    Bot & How about Arrival, starring Amy Adams. If you want, you can say who else is in it.\\
    User & Who else is in it? [intent = GetNextActorIntent] \\
    Bot & The character of Ian Donnelly was played by Jeremy Renner.\\
    User & Is it any good? [intent = GetRatingIntent] \\
    Bot & It is rated 8.4 on I.M.D.b., based on about 70000 votes.\\
    User & Thank you [intent = AMAZON.StopIntent] \\
    Bot & Thank you for using MovieBot.\\
    \bottomrule
  \end{tabularx}
  \egroup
  \end{adjustbox}
\end{table} 
 
Providing the sampled text directly to the MovieBot NLU model would result in practically zero errors as the sample utterances are used in training the NLU model via the ASK portal. Therefore, we take the approach of generating a spoken audio signal from the text, injecting noise into the audio and then passing the audio to ASR and NLU models of the MovieBot Skill. This process simulates the real-world sources of error such as background noise in a person's home. Concretely, we synthesize an audio signal with the text-to-speech (TTS) service provided by Amazon Polly \cite{polly} and contaminate the clean TTS output using PyAcoustics \cite{PyAcoustics} to achieve a desired signal-to-noise ratio (SNR). Contamination is performed by filtering a white noise signal using a Fast Fourier transform applied to the original TTS output. Given our choice of architecture, any noise injection strategy may be developed without impacting the rest of the experimental setup. For example, in the future we plan to explore the impact of using different kinds of noise such as background noise, speaker-dependent speech variations, or multiple concurrent speakers. In this particular set of experiments, we tune the injected white noise to a level that causes the ASR WER of the simulator to match that of MovieBot's deployed ASR model.

This approach to simulating users has a number of benefits. By predicting intents first, the model becomes highly accurate and data efficient at the intent level due to the smaller number of intents compared to words in the language. By using RNNs, our simulator learns long-term dependencies between dialog context and intents from data to predict consistent user behavior. Additionally, using handcrafted templates ensures that the generated utterances are of high quality, while the TTS component with synthesized noise allows us to simulate real-world noisy environments.

\subsection{Implementation and Evaluation}

We used a corpus of $966$ MovieBot dialogs consisting of $6349$ user turns to train the intent generation model with and without agent responses. The corpus is collected from real user interactions with MovieBot and annotated at the utterance level. For example, a user request such as "How old is Tom Hanks?" is annotated as intent=\textit{GetAgeIntent} and slot \textit{PersonName}=\textit{Tom Hanks}. For intents and slot types, we added an embedding layer initialized with small random values. We used a single layer network followed by softmax, and for the hidden units we used two variants: vanilla RNN and Gated Recurrent Unit (GRU) \cite{gru}. The output of the network is a sequence of vectors, each with total number of intent/slot pair elements. The optimal parameters were found using grid search (Appendix A). For all models we set the learning rate to $0.001$, and used Google's trained word2vec model for encoding agent responses. As a baseline, we used a bigram model. Note that this model only takes the intent and slot type inputs. With this setting, we found that the vanilla RNN models have the lowest perplexity \cite{serban} scores across various trials (Table \ref{quant_results}). 

\begin{table}[b]
\small
\centering
\caption{Perplexity results for the Bigram, RNN, and GRU models. Note that a lower value indicates better performance.}
\label{quant_results}
\begin{adjustbox}{max width=\textwidth}
\begin{tabular}{l|c|c|c}
  \hline
  \multicolumn{1}{c|}{Model} & Bigram & RNN & GRU \\
  \hline
  Intent and Slot Only & 16.270 & \textbf{7.675} & 7.769 \\
  Intent and Slot + Bot Response & $-$  & \textbf{7.274} & 7.892 \\
  \hline
\end{tabular}
\end{adjustbox}
\end{table}

We also conducted a user study to evaluate if participants could distinguish simulator generated dialogs from real human dialogs. The participants were each presented $10$ dialogs via a web interface and asked to determine if each dialog was generated by a human or a simulated user. Optionally, the participants could also provide feedback as to why they made their decision. For each participant, five dialogs were sampled from the annotated MovieBot data, and the first utterance of each was used to generate five dialogs with the simulator. We used two models for the simulated data, namely, the bigram model and the vanilla RNN model (see Appendix B for example of simulated dialogs). The difference between the true positive rate (human dialog correctly identified as human) and false positive rate (simulated user incorrectly identified as human) is of particular interest: more individuals identifying simulated user dialogs as human indicates better user simulation.

\begin{table}[t]
\small
  \centering
  \caption{Results of the user study for distinguishing between real and simulated dialogs. The difference between identifying human as human and simulator as human is small.}
  \label{user-study-table}
  \begin{tabular}{l|cl|cl}
    \toprule
         & \multicolumn{2}{c|}{All Participants} & \multicolumn{2}{c}{Participants Who}  \\
         & & & \multicolumn{2}{c}{Left Feedback} \\
    \midrule
    Number of Subjects & \multicolumn{2}{c|}{31}  & \multicolumn{2}{c}{13}     \\
   Total Number of Dialogs &  \multicolumn{2}{c|}{309}  & \multicolumn{2}{c}{129}     \\
    Accuracy (Human/RNN + Bigram Models) & \multicolumn{2}{c|}{50.49\%} & \multicolumn{2}{c}{58.14\%} \\
    Precision (Human/RNN + Bigram Models)  & \multicolumn{2}{c|}{50.29\%}  & \multicolumn{2}{c}{57.14\%}  \\
    Recall (Human/RNN + Bigram Models)      & \multicolumn{2}{c|}{57.14\%}   & \multicolumn{2}{c}{62.50\%}  \\
    F1-Score (Human/RNN + Bigram Models)  & \multicolumn{2}{c|}{53.50\%} & \multicolumn{2}{c}{59.70\%} \\
    \midrule
    Human Identified as Human & 57.14\% & 154 dialogs & 62.50\% & 64 dialogs\\
    Simulator Identified as Human (RNN + Bigram Models)  & 56.13\% & 155 dialogs & 46.15\% & 65 dialogs\\
    Simulator Identified as Human (RNN Model) & 59.21\% & 76 dialogs & 58.82\% & 34 dialogs \\
    Simulator Identified as Human (Bigram Model) & 53.16\% & 79 dialogs & 32.26\% & 31 dialogs \\
    \bottomrule
  \end{tabular}
\end{table}

In total, $31$ individuals from the Amazon Alexa team participated in the study and evaluated $309$ dialogs. The results are reported in Table \ref{user-study-table}. Overall results indicate that the difference between identifying human as human ($57.14\%$) and simulator as human ($56.13\%$) is not statistically significant (p-value = $0.858$). Additionally, the RNN model is performing slightly better than the bigram model. Upon further inspection, we found a weak correlation (Pearson's correlation coefficient: $0.38$, p-value: $0.036$) between how participants rated a dialog and whether or not they left feedback. The results for the $13$ participants who left feedback is reported separately in the table, clearly showing the RNN ($58.82\%$ of simulated users identified as human) significantly (p-value = $0.033$) outperforms the bigram model ($32.26\%$ of simulated users identified as human). We hypothesize that this is due to bigram dialogs being longer on average, and the participants who left feedback read the dialogs more carefully. Criteria for selecting a dialog as a real human as indicated in the participant feedback included: the expression of emotions (e.g., the user gets mad when the bot doesn't answer correctly), logical progression of the dialog, specific goals that humans have regardless of the dialog system's response, and naturalness of the dialogs. On the other hand, top criteria for identifying simulated users included: asking a question without context, referring to non-existent entities in the dialog, random responses by the user, and the user not getting frustrated. These comments illustrate that better context tracking and use of sentiments would further improve user simulation. Nevertheless, given the small difference between true positive and false positive results, the user simulator can sufficiently replace real users for dialog policy learning.

\section{Dialog Policy Learning}

We used the proposed user simulator to train dialog policies through deep RL with the goal of conversational error recovery -- that is, recovering from ASR/NLU errors through necessary clarifications. The components of the dialog policy learning problem are as follows:

\begin{itemize}
   \item \textit{S}: The state is composed of 1) the hypothesis intent and slot, 2) the ASR and NLU intent and slot confidence scores, and 3) the previous bot action. 
   \item \textit{A}: We constrain the action space to three actions that are critical for conversational error recovery: \textit{execute}, \textit{confirm}, and \textit{elicit}. \textit{execute} accesses an IMDb database and answers the user's question. \textit{confirm}  and \textit{elicit} are used to recover from ASR/NLU errors. \textit{confirm} clarifies the intent and/or the slot with the user (e.g., \textit{"Do you want movies directed by Christopher Nolan?"}), whereas \textit{elicit} asks the user for missing information (e.g., \textit{"Which movie are you talking about?"}).
   \item \textit{R}: Table \ref{reward_function} summarizes the reward function. Here, \textit{ref} indicates the reference intent and slot as generated by the user simulator, and \textit{hyp} refers to what the bot understands. If $hyp \neq ref$, this indicates that an NLU error was induced by added noise to the audio signal. The environment gives a large positive reward for executing correctly, a large negative reward for executing incorrectly, and smaller negative rewards for confirm and elicit. A successful execution is when the reference and hypothesis intent and slots are equal. Clarification, although sometimes necessary, can frustrate users, hence the small negative rewards for clarifying actions. Additionally, if the user leaves the conversation after a successful execution, the bot receives a small positive reward, but if the user leaves after an incorrect execution, the bot receives a medium negative reward. The reason we give a small positive reward to termination after a successful execution is that in the absence of an explicit goal we assume the user received the requested information and decided to terminate the dialog. All specific values were assigned through empircal analysis of user interactions with the simulator.
\end{itemize}

\begin{table}[t]
\small
  \centering
  \caption{Reward Function}
  \label{reward_function}
  \begin{tabular}{l|c|c}
    \toprule
    \multicolumn{1}{c|}{\textit{a}}  & \textit{s} & \textit{r} \\
    \midrule
	execute(\textit{hyp}) & $hyp = ref$ & $+10$ \\
	execute(\textit{hyp}) & $hyp \neq ref$ & $-12$ \\
	confirm & $*$ & $-3$ \\
	elicit & $*$ & $-6$ \\
	terminate (after correct execute) & $*$ & $+1$ \\
	terminate (after incorrect execute) & $*$ & $-5$ \\
    \bottomrule
  \end{tabular}
\end{table}

For the deep RL experiments we used DQN and Dueling DDQN \cite{dueling_ddqn}, with a fully-connected Multi-Layer Perceptron (MLP) to represent the deep Q-network. The input of the network is the concatenated embedding vector of intent and slot, the vector of confidence scores, and previous bot action. Additionally, we tuned a window size to include previous dialog turns as input. The hidden layers use a rectifier nonlinearity, and the output layer is a fully connected layer with linear activation function and a single output for each valid action. We trained the agents using an $\epsilon$-greedy policy with $\epsilon$ decreasing linearly from 1 to $0.1$ over 100,000 steps. We ran each method $30$ times for 150,000 steps, and in each run, after every 10,000 steps, we sampled $30$ dialog episodes with no exploration to evaluate performance. All methods used the same set of random seeds and the best parameters were empirically found for each method using Hyperopt \cite{hyperopt}. Additionally, we experimented with two fixed policies: 1) \textit{execute} only and 2) \textit{execute} with \textit{confirm} and \textit{elicit}. For the second fixed policy, we learned global ASR and NLU confidence score thresholds for MovieBot by optimizing the simulated agent's average return (Table \ref{fixed_thresholds}).

\begin{table}[ht]
\small
  \centering
  \caption{Fixed Policy (MovieBot - Execute/Confirm/Elicit)}
  \label{fixed_thresholds}
  \begin{tabular}{l|l}
    \toprule
    \multicolumn{1}{c|}{\textit{a}}  & \multicolumn{1}{c}{\textit{s}}  \\
    \midrule
	execute(\textit{hyp}) & $score_{ASR} \geq 0.34$ AND $score_{NLU} \geq 0.56$ \\
	elicit & $score_{ASR} < 0.06$ \\
	confirm & $otherwise$ \\
    \bottomrule
  \end{tabular}
\end{table}

\begin{figure}[t]
  \centering
  \includegraphics[width=0.70\textwidth]{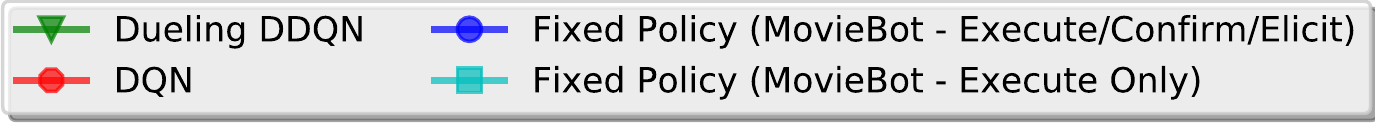}
    \begin{subfigure}{0.4\linewidth}
  		\includegraphics[width=1\textwidth]{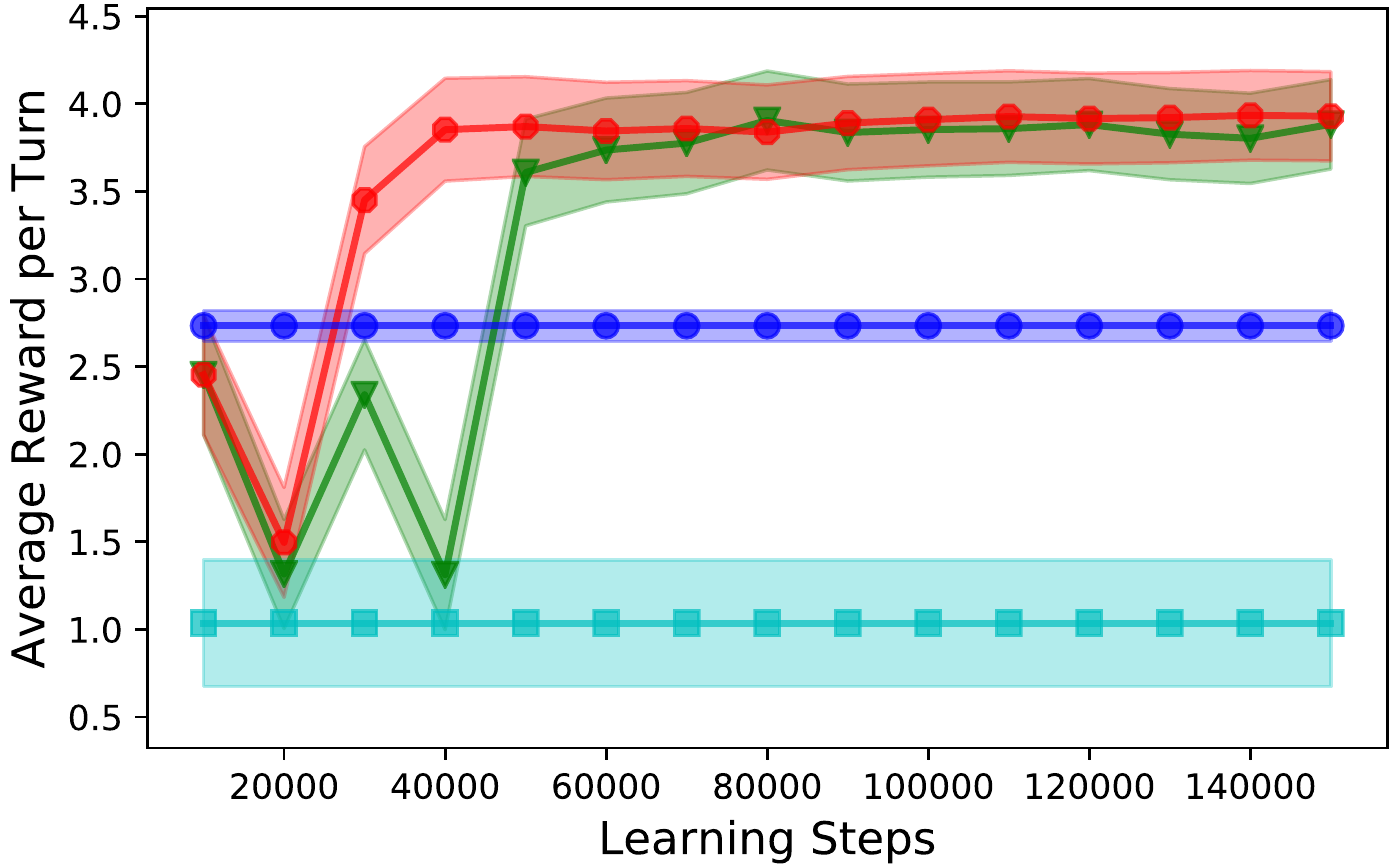}
  	\caption{}
  	\label{average_reward}
    \end{subfigure}
    \hspace{1mm}  
  	\begin{subfigure}{0.4\linewidth}
  		\includegraphics[width=1\textwidth]{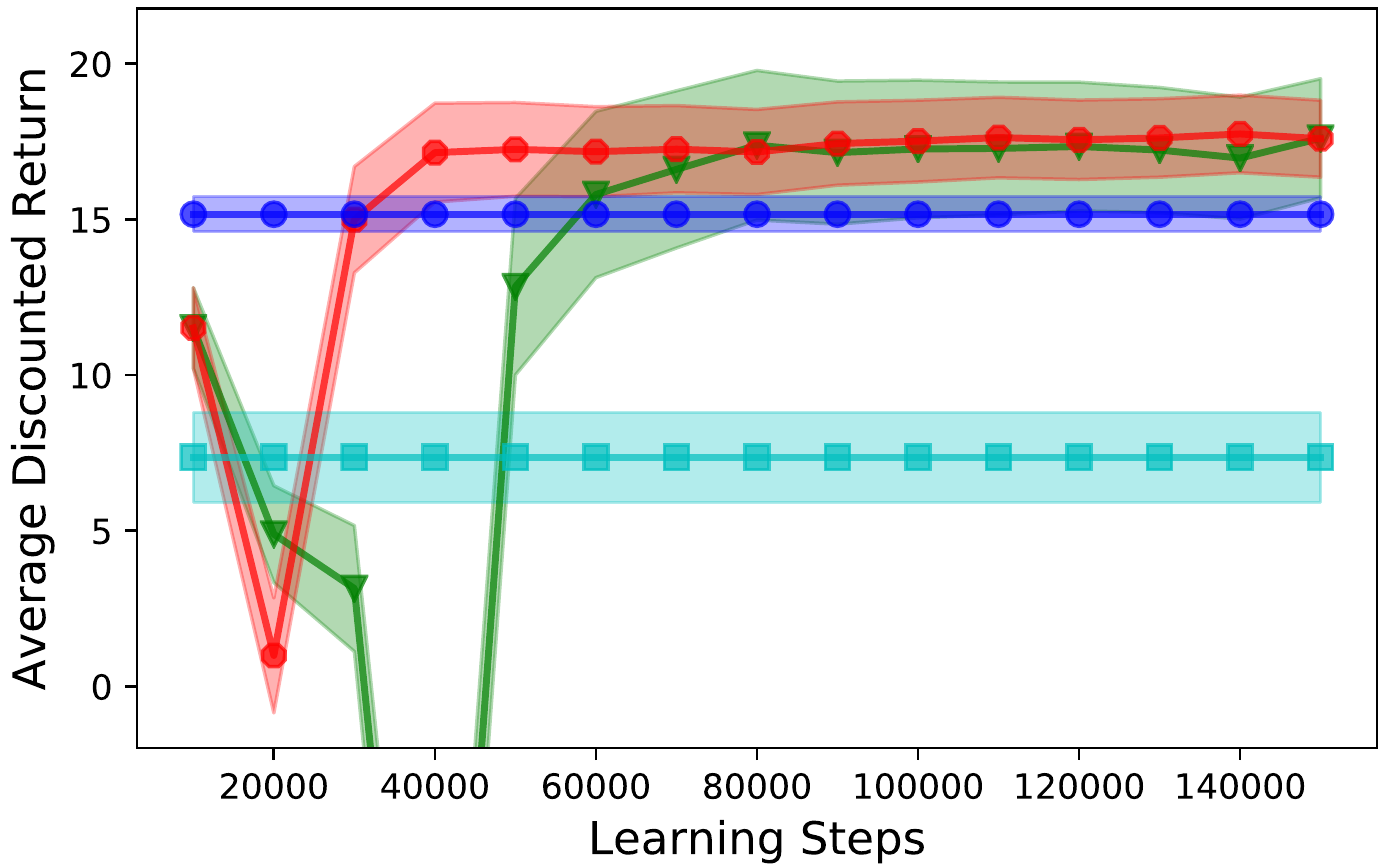}
  	\caption{}
  	\label{average_return}
    \end{subfigure}
    \hspace{1mm}
  	\begin{subfigure}{0.4\linewidth}
  		\includegraphics[width=1\textwidth]{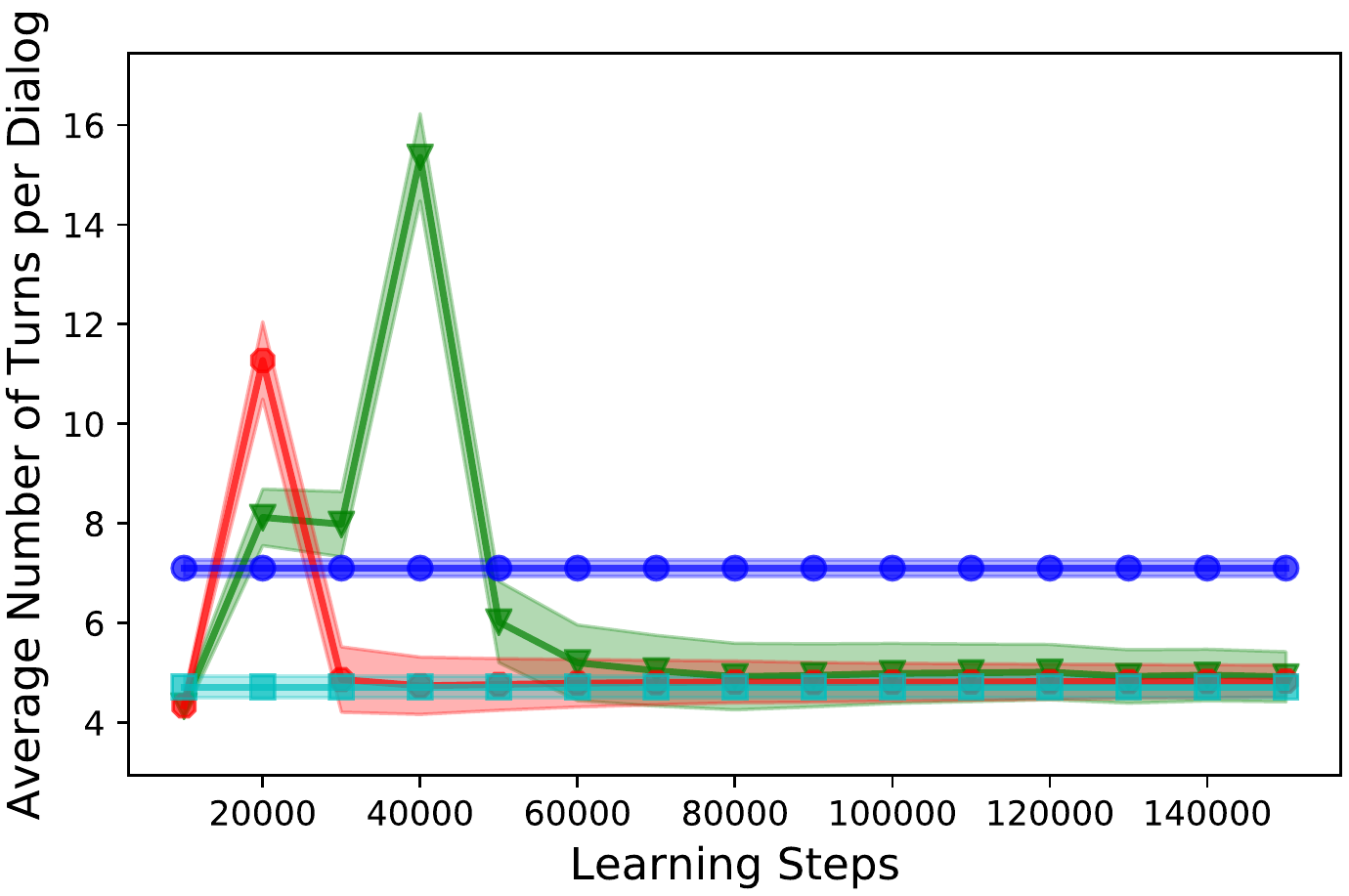}
  	\caption{}
  	\label{average_turn}
    \end{subfigure}
    \hspace{1mm}
  	\begin{subfigure}{0.4\linewidth}
  		\includegraphics[width=1\textwidth]{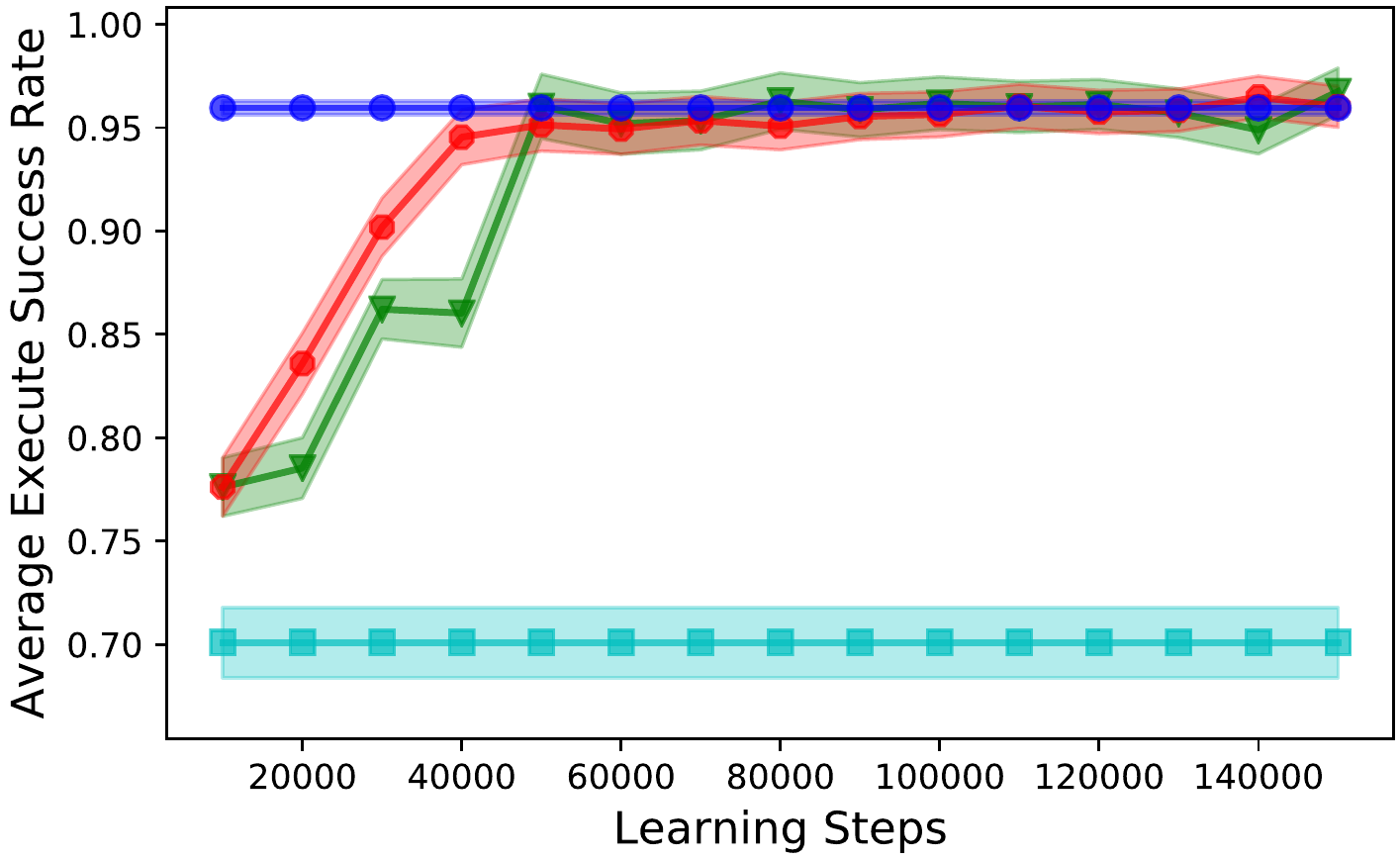}
  	\caption{}
  	\label{average_success}
    \end{subfigure}
  \caption{Results during policy learning process. a) Average reward per turn, b) average discounted return, c) average number of turns per dialog, and d) execution success rate. Shaded areas highlight the standard error of the mean. RL approaches achieve our goal of recovering from ASR/NLU errors without frustrating the user and result in around $40\%$ relative gain in average rewards per turn over the best rule-based method, with the same success rate and 2 fewer turns per dialog.}
  \label{learning_plots}
\end{figure}

\subsection{Experimental Results}

For our experiments, we varied the amount of noise such that we match the ASR WER of the MovieBot corpus. This was achieved when SNR was set to $2.8$. Figure \ref{learning_plots} shows the simulation results (see Appendix C for learned parameters). The Y-axis in the four subfigures is the average reward per turn, the average discounted return, the average number of turns per dialog, and the execution success rate of each technique, respectively. The X-axis in all four figures shows the number of learning steps. Shaded areas highlight the standard error of the mean. RL approaches have the same dialog episodes with the user with fewer clarifying questions, and as such result in around $40\%$ relative gain in average rewards per turn over the best rule-based method (Figure \ref{average_reward}). Furthermore, they achieve the same success rate (Figure \ref{average_success}) with 2 fewer turns per dialog on average (Figure \ref{average_turn}).

\begin{figure}[b!]
  \centering
  		\includegraphics[width=0.5\textwidth]{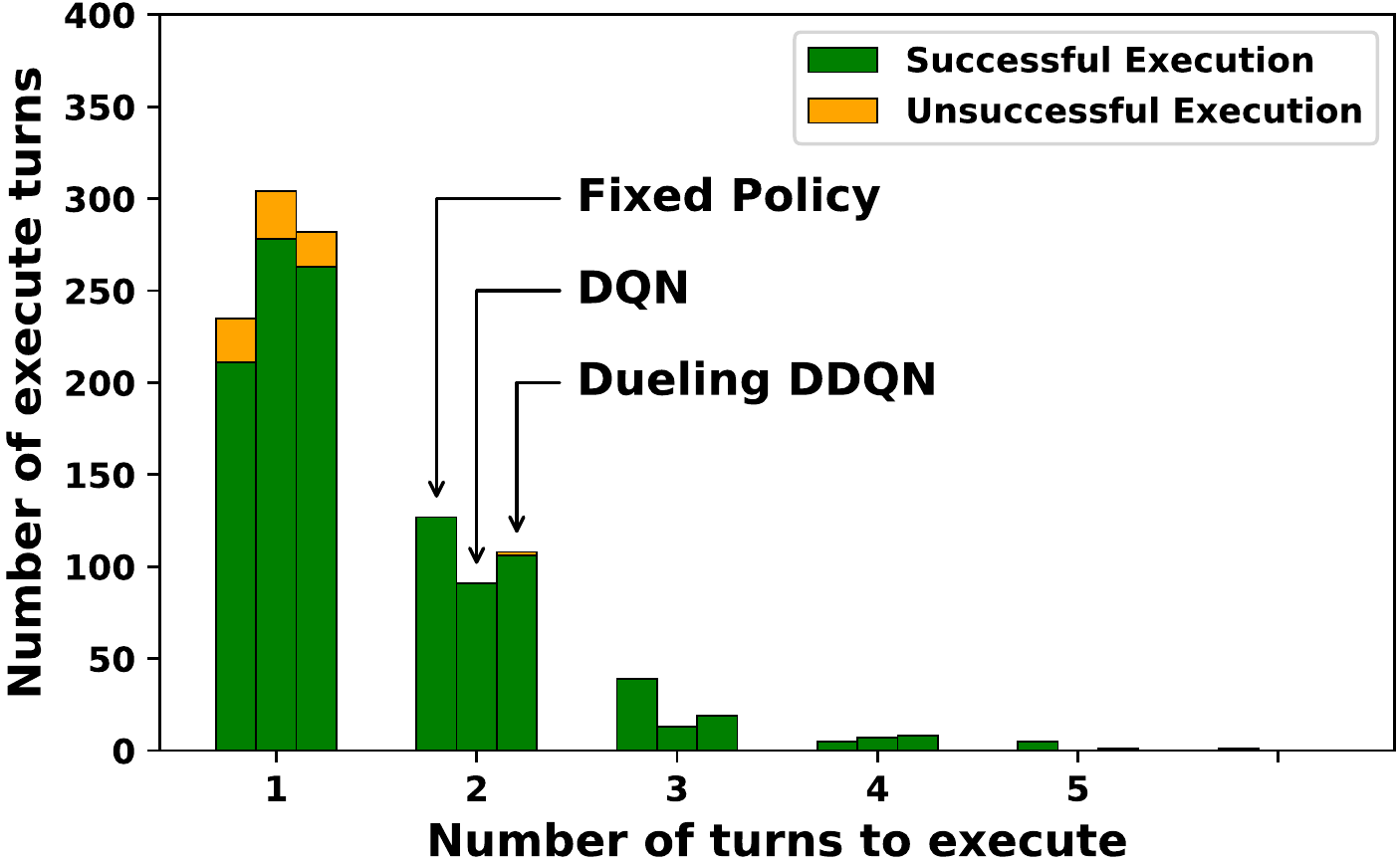}
  \caption{The number of turns needed in order to execute for Fixed Policy (Execute/Confirm/Elicit), DQN and Dueling DDQN when testing on the learned policies. RL approaches learn to ask fewer clarifying questions while achieving the same success rate.}
  \label{num_to_execute}
\end{figure}

To better understand the advantage of RL approaches, Figure \ref{num_to_execute} illustrates what a user experiences in order to obtain information in terms of number of turns needed to execute for each agent, where $1$ means execute without any clarifying question. This figure illustrates that RL approaches learn to ask fewer clarifying questions while achieving a similar success rate. In other words, the percentage of unsuccessful executions stays the same with $6.2\%$ for the Fixed Policy, $7.2\%$ $(\pm 1.3\%)$ for DQN, and $5.0\%$ $(\pm 1.7\%)$ for Dueling DDQN, with 0.2 less turns per execution. 

To gain insight into the learned policy's properties, we conducted a qualitative analysis to compare $100$ dialogs generated by Dueling DDQN and Fixed Policy agents. To achieve a fair comparison, we fixed the random number generator so that the user simulator generated identical sequences of intents and utterances to interact with both agents. Overall, $28\%$ of the dialogs were identical. Considering the remaining dialogs, two main reasons for the differences can be identified. First, the DDQN agent can decide not to clarify based on the combination of specific intent and confidence scores. Table \ref{learning_qualitative_analysis_example_elicit} shows one dialog turn between the simulator and the two agents, along with the ground truth of the simulated utterance and intent (first column of the table), the hypothesis from MovieBot's ASR/NLU component (second column), and the confidence scores of the hypotheses with a range between 0 and 1 (third column). In this example, since the ASR score is relatively low ($0.239$), the Fixed Policy agent decides to ask a clarifying question (denoted as \texttt{<confim>}). However, the low ASR score comes from the misrecognition of words which have small effect on the recognition of the intent as indicated by the high NLU confidence score. As such, the DDQN agent learns that the ASR confidence score is less important in this case and the agent executes the command and provides the requested information. Secondly, the DDQN agent learns that elicitation brings more information than confirmation since the user simulator repeats the original utterance. $82\%$ of the clarifying questions asked by the DDQN agent are elicitation, as opposed to $61\%$ for the Fixed Policy.

\begin{table}[t]
\small
  \centering
  \caption{One dialog turn between a simulated user and Fixed Policy (MovieBot - Execute/Confirm/Elicit) agent, and Dueling DDQN agent, along with the ground truth of the utterance and intent (first column of the table), the hypothesis from the ASR/NLU components (second column), and the confidence scores of the hypotheses with a range between 0 to 1 (third column). The DDQN agent can decide not to clarify based on the combination of specific intent and confidence scores.}
  \label{learning_qualitative_analysis_example_elicit}
  \begin{tabularx}{1.0\textwidth}{l|X}
    \toprule
    Fixed Policy & USER: recommend a popular movie \\
    & BOT: Do you want popular movies? <confirm> \\
    \midrule
    Dueling DDQN & USER: recommend a popular movie \\
    & BOT: A popular movie is It, a Drama Horror film, starring Bill Skarsgård. <execute> \\
     \bottomrule
 \end{tabularx}
  \begin{tabular}{l|l|c}
  \toprule
	\multicolumn{1}{c|}{reference} & \multicolumn{1}{c|}{hypothesis} & \multicolumn{1}{c}{confidence score} \\
    \midrule
	\multicolumn{1}{l|}{recommend a popular movie} & \multicolumn{1}{l|}{popular movies} & \multicolumn{1}{c}{0.239}\\
	\midrule
	\multicolumn{1}{l|}{GetPopularMoviesIntent} & \multicolumn{1}{l|}{GetPopularMoviesIntent} & \multicolumn{1}{c}{0.769} \\
    \bottomrule
  \end{tabular}
\end{table}

\section{Conclusion}

In this paper, we demonstrate a method for learning robust dialog policies despite upstream NLU/ASR errors caused by noisy environments. The user simulator presented here is able to mimic realistic conversations, nearly indistinguishable from human dialogs, in the context of a real-world application -- the MovieBot Alexa Skill. While other simulators inject noise via stochastic processes at the intent or slot level~\cite{simulator_vivian}, our simulator is able to leverage audio signals and white noise to train policies in a production-like environment. Using this simulator, we apply deep RL to learn dialog policies which achieve the same success rate with fewer dialog turns compared to fixed rule-based policies in noisy environments. Based on feedback from the participants in our user study, an interesting area of future work is to use sentiment to further improve the user simulator. That is, user sentiment can be sampled to generate polarized target utterances to simulate user frustration or satisfaction. Furthermore, we plan to investigate other dimensions of dialog to improve our simulator and RL-based dialog policy learning methods. These include per-interaction user satisfaction and efficiency of interactions to model the reward function. Lastly, we plan to explore the impact of using different kinds of noise on the learned policies.

\section*{Acknowledgments}

We thank Jason Pazis and other members of the Conversational AI group at Amazon for helpful discussions and comments. We also thank our anonymous reviewers
for insightful feedback. 

\section*{Appendix}
\renewcommand{\thesubsection}{\Alph{subsection}}
\subsection{Intent Generation Hyper-parameters}

\begin{table}[h]
  \centering
  \label{rnn_params}
  \begin{adjustbox}{max width=\textwidth}
  \begin{tabular}{l|c|c|c|c}
    \toprule
     & RNN Model & RNN Model + Bot Response & GRU Model & GRU Model + Bot Response \\
    \midrule
	Number of Hidden Nodes per Layer & 30 & 90 & 10 & 10  \\
	Intent and Slot Embedding Size & 50 & 50 & 70 & 70  \\
	Number of Epochs &  100 & 30 & 10 & 10 \\
    \bottomrule
  \end{tabular}
  \end{adjustbox}
\end{table}

\subsection{Example of simulated dialogs identified as human dialogs}

\begin{table}[H]
 \small
  \centering
  \label{simulator-example}
  \centering
  \begin{adjustbox}{max width=\textwidth}
  \bgroup
  \def\arraystretch{1.1}
  \begin{tabularx}{\linewidth}{lX|lX}
    \toprule
    \multicolumn{2}{c}{\textbf{Bigram Simulator Dialog}} & \multicolumn{2}{|c}{\textbf{RNN Simulator Dialog}} \\
    \midrule
    User & ummm comedy & User & ummm comedy \\
    Bot & I'd recommend Kingsman: The Golden Circle, a comedy action film starring Taron Egerton & Bot & I think you might like Cars 3, a comedy adventure film, starring Owen Wilson.\\
    User & Look for something else. & User & How long is the movie? \\
    Bot & I really liked Moana, a comedy adventure film starring Auli'i Cravalho. & Bot & The movie duration is 1 hour and 42 minutes.\\
    User& What's my favorite genre? & User & Who wrote the soundtrack? \\
    Bot& Sci-fi, because it has been in 25 percent of your genre-related requests. & Bot & I don't know how to handle that yet. You can also say, can children watch this movie. \\
    User & Give me a blooper from the movie inception. & User & What is a popular romantic comedy movie? \\
    Bot & Here is an interesting inception goof. *gives goof* & Bot & I'd recommend Blind, a romance drama film starring Demi Moore. \\
    User & How does John Wick rate? & User & Tell me about it. \\
    Bot&  John Wick has a rating of 7.3 on IMDb., based on about 300000 votes. & Bot & Here is a summary of blind from IMDb. *gives plot summary* \\
    User & Cancel & User & That is enough. \\
    Bot & Thank you for using MovieBot. & Bot & Thank you for using MovieBot. \\
    \bottomrule
  \end{tabularx}
  \egroup
  \end{adjustbox}
\end{table} 

\subsection{Learning parameters for DQN and Dueling DDQN for SNR of 2.8}

\begin{table}[h]
\small
  \centering
  \label{parameters}
  \begin{tabular}{l|c|c}
    \toprule
    \multicolumn{1}{c|}{}  & \multicolumn{1}{c|}{DQN} & \multicolumn{1}{c}{Dueling DDQN} \\
    \midrule
	Number of Hidden Layers & 2 & 3 \\
	Number of Hidden Nodes per Layer & 32 & 128 \\
	Embedding Size & 5 & 30 \\
	Dropout & 0.5 &  0.0 \\
	Learning Rate & 0.0001 & 0.00001 \\	
	Experience Replay Size & 10,000  & 15,000 \\
	Window Size & 2 & 8 \\
	Discount Factor & 0.97 & 0.97 \\
	Target Model Update Interval & 12,000 & 8,000 \\
    \bottomrule
  \end{tabular}
\end{table}

\small

\bibliographystyle{named}

\end{document}